\documentclass[final]{cvpr}

\usepackage{times}
\usepackage{epsfig}
\usepackage{graphicx}
\usepackage{amsmath}
\usepackage{amssymb}
\usepackage{amsfonts}
\usepackage{color}
\usepackage{multirow}
\usepackage{booktabs}

\newcommand{\tabincell}[2]{\begin{tabular}{@{}#1@{}}#2\end{tabular}}

\usepackage[pagebackref=true,breaklinks=true,colorlinks,bookmarks=false]{hyperref}

\def\cvprPaperID{1024} 
\def\confYear{CVPR 2021}

\begin{document}

\title{View-Guided Point Cloud Completion}

\author{Xuancheng Zhang\textsuperscript{\rm 1}, Yutong Feng\textsuperscript{\rm 1}, Siqi Li\textsuperscript{\rm 1} Changqing Zou\textsuperscript{\rm 3}, Hai Wan\textsuperscript{\rm 1},\\ Xibin Zhao\textsuperscript{\rm 1} \thanks{Corresponding authors}, Yandong Guo\textsuperscript{\rm 4}, Yue Gao\textsuperscript{\rm 1, 2} \footnotemark[1]\\  
\textsuperscript{\rm 1}BNRist, KLISS, School of Software, Tsinghua University, China\\ \textsuperscript{\rm 2}THUICBS, Tsinghua University,
\textsuperscript{\rm 3}Huawei Technologies Canada Co., Ltd, \textsuperscript{\rm 4}OPPO Research Institute\\
\tt\small \{zxc19,fyt19,lsq19\}@mails.tsinghua.edu.cn, aaronzou1125@gmail.com \\ \tt\small\{wanhai,zxb,gaoyue\}@tsinghua.edu.cn, yandong.guo@live.com}


\maketitle

\begin{abstract}
This paper presents a view-guided solution for the task of point cloud completion. 
Unlike most existing methods directly inferring the missing points using shape priors, we address this task by introducing ViPC (view-guided point cloud completion) that takes the missing crucial global structure information from an extra single-view image. By leveraging a framework 
that sequentially performs effective cross-modality and cross-level fusions, our method achieves 
significantly superior results over typical existing solutions on a new large-scale dataset we collect for the view-guided point cloud completion task. 

\end{abstract}


\vspace{-0.5cm}
\section{Introduction}
Point cloud has attracted increasing research interest due to its wide range of applications in various fields such as auto-driving \cite{kato2018autoware0}, robotics \cite{pomerleau2015review}, geography \cite{rieg2014data}, phenomics \cite{lin2015lidar}, and archaeology \cite{brutto2012computer}. In practice, the point cloud's data quality directly acquired by the depth scanning devices can be affected by many factors such as occlusions between objects and low scanning precision, which may lead to the poor-quality point cloud, \textit{e.g.}, incomplete, sparse, and noisy point cloud.   

Existing methods, mainly including point cloud completion \cite{yuan2018pcn}, denoising \cite{castillo2013point},  and super-resolution (up-sampling) \cite{yifan2019patch}, have been proposed for the task of point cloud enhancement.
Early methods generated enhanced point cloud by mainly using shape prior information \cite{kazhdan2013screened} or hand-crafted geometric regularities \cite{thrun2005shape}. In recent years, data-driven methods, especially deep learning techniques like PointNet \cite{qi2017pointnet} and DGCNN \cite{wang2018dynamic}, have made significant progress on this problem. Compared to traditional methods, these deep learning based methods have demonstrated significant advantages in processing objects with irregular structure and geometry.

\begin{figure}[t]
\begin{center}
  \includegraphics[width=1\linewidth]{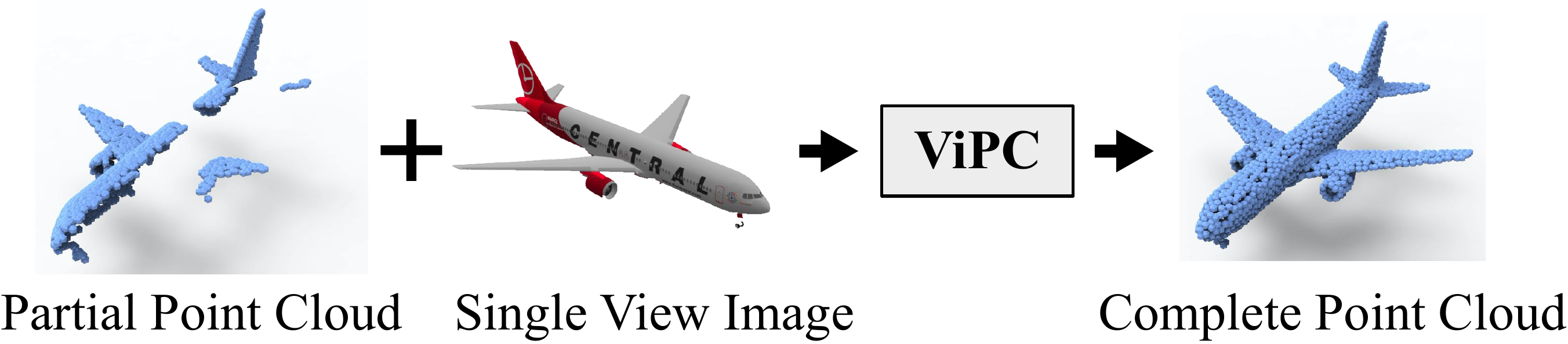}
\end{center}
   \caption{ViPC is a new approach completing a partial point cloud by leveraging the complementary information from an extra single-view image.}
\label{ViPC}
\vspace{-0.5cm}
\end{figure}

In this paper, we focus on the following point completion task: the input point cloud is incomplete but {with limited noise}, while our method outputs a complete point cloud. Studying this problem addresses a common problem in real-world 3D data acquisition where a 3D scanner with a RGB camera is occluded by other objects in the environment.   
The most recent solutions to this problem are data-driven, leveraging an encoder-decoder architecture~\cite{dai2017shape, yuan2018pcn, Park_2019_CVPR,xie2020grnet}. In those methods, an encoder transfers the incomplete input point cloud into the feature space, and then a decoder reconstructs a complete point cloud by transferring the features back to Euclidean space. 
The whole network works as a parameterized model by learning a mapping between the two latent spaces of incomplete and complete point cloud. 
In the cases where there is a large degree of incompleteness in the input point cloud, learning this mapping with only the single-modality point cloud data is challenging because of the following factors: 1) there is a great uncertainty in inferring the missing points due to the limited amount of information available, 2) point cloud is of an unstructured data, together with inherent sparseness, it is difficult to determine whether a blank 3D space is caused by inherent spareness or incompleteness. 

In this paper, we seek a more applicable solution to the point cloud completion task. Specifically, we address the task with the help of the image modality and propose a view-guided point completion framework (ViPC) as illustrated in Figure.~\ref{ViPC}. This setting of sensor fusion is increasingly common as the hardware cost decreases (e.g., the Intel Real Sense D455 and Microsoft Kinect devices). The key challenge of solving this problem is how to effectively fuse the information of pose and local details provided by the partial point cloud and the global structure information provided by the single-view image.
This is not trivial because it involves a two-dimensional challenge: ``cross-modality" (the information is from both image and point cloud modalities) and ``cross-level" (local details and global structure are information from different levels). 
We address the problem by a three-stage framework which first
address the cross-modality challenge and then the cross-level challenge.
Specifically, the cross-modality challenge is addressed by reconstructing a coarse point cloud from the single-view image and transferring all the information required by the completion to the identical point cloud domain. The cross-level challenge is addressed by a differential refinement strategy empowered by a network
called ``Dynamic Offset Predictor" which can deferentially refine the points in a coarse point cloud: performing a light refinement for low-quality points while a heavy refinement for high-quality points. 

To better investigate the problem, we have built a large-scale dataset called ShapeNet-ViPC on
existing ShapeNet dataset~\cite{chang2015shapenet}. Our dataset includes 38,328 objects from 13 categories. Each object has 24 sets of ground-truth data consisting of two incomplete point clouds produced under two typical data acquisition scenarios, a view-aligned image and a complete ground-truth point cloud. Extensive evaluations on ShapeNet-ViPC demonstrate that the proposed approach can achieve significantly superior results than existing state-of-the-art approaches.

In summary, the main contributions of our methods are threefold:
\begin{enumerate}
\item We propose a new solution for point cloud completion, in which an extra single-view image explicitly provides the crucial global structural prior information for completion. 
\item {We design a new general deep network for point cloud refinement which can deferentially refine the points in a point cloud.}
%
\item We build a large-scale dataset for the point cloud completion task on the ShapeNet dataset. This dataset simulates point cloud defects caused by various kinds of occlusions.
It could be used as a benchmark for future research of point cloud completion.
\end{enumerate}

\begin{figure*}[t]
\begin{center}
  \includegraphics[width=1\linewidth]{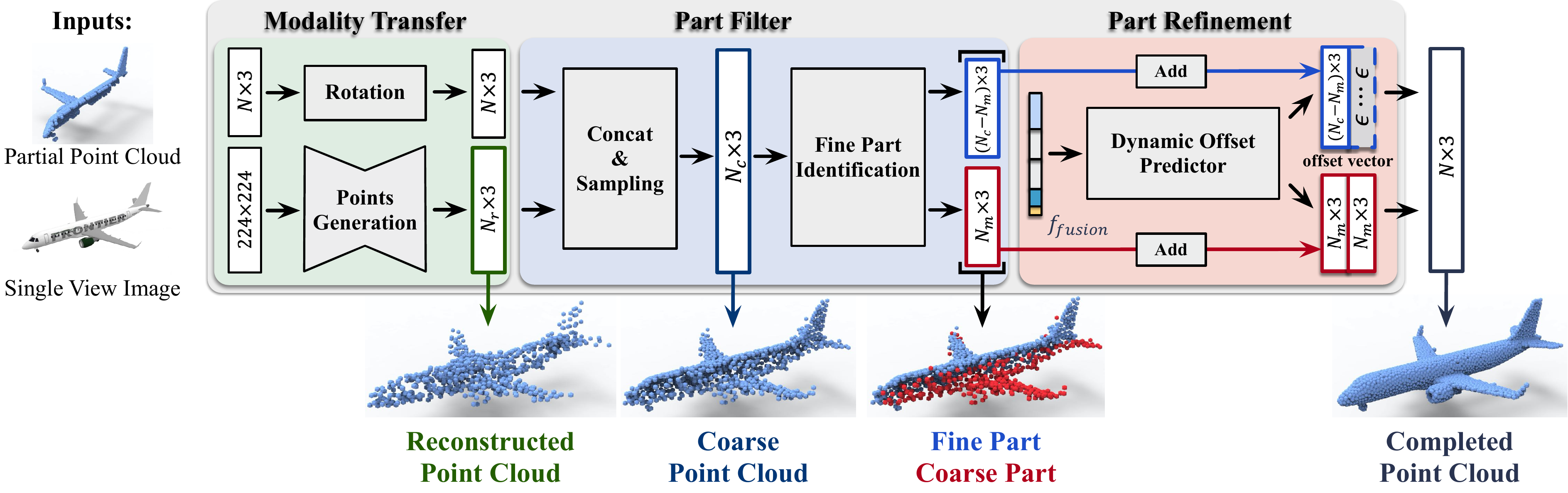}
\end{center}
\vspace{-0.3cm}
   \caption{
   {Architecture of the proposed ViPC; see the Pipeline section in the text for details.}}
\label{fig:pipeline}
\vspace{-0.3cm}
\end{figure*}

\section{Related Works}


Existing point cloud completion methods can be generally classified into three types of approaches: geometry-based, alignment-based and learning-based. \\

\textbf{Geometry-based Methods.}
Geometry-based methods predict the invisible shape part from the observed shape part directly by prior geometric assumptions~\cite{6849979, 8672183}. More specifically, some methods fill the surface holes locally by generating smooth interpolations \cite{berger2014state, thanh2016field}, such as Laplacian smoothing \cite{nealen2006laplacian} and Poisson surface reconstruction \cite{kazhdan2013screened}. Other methods detect the regularities in model structures and repeat them to predict missing data based on the identified symmetry axes \cite{mitra2006partial, pauly2008discovering, thrun2005shape}, \. These methods infer the missing data directly from the observed region and show impressive results. However, they need the hand-crafted geometric regularities that are predefined for specific kinds of models and only applied to models with a small degree of incompleteness.

\textbf{Alignment-based Methods.} Alignment-based methods retrieve identical models similar to the target object in a shape database, then align the input with temple models and complete the missing region. Some methods retrieve 3D shapes directly, such as the whole model \cite{pauly2005example} or part of them \cite{kalogerakis2012probabilistic, kim2013learning}. Other methods use synthesized models after deformation \cite{rock2015completing} or non-3D geometric primitives such as planes and quadrics \cite{chauve2010robust,schnabel2009completion,yin2014morfit} in place of 3D shapes in the database. These methods are applicable to many different types of models and can be applied to varying degrees of incompleteness, but they require expensive cost during inference optimization and database construction; also they are sensitive to noise. 

\textbf{Learning-based Methods.} Learning-based methods construct a parameterized model to learn a mapping between the two feature spaces of the incomplete and complete point cloud of a shape. Most of them are encoder-decoder based neural networks. As for shape representation, most existing models use voxels to represent a shape~\cite{dai2017shape,han2017high}, which are intuitive and convenient for 3D convolution. In order to preserve more geometric information (i.e., local geometric details) in the completed point cloud, several models perform the operation on point sets directly \cite{yuan2018pcn,yang2018foldingnet,tchapmi2019topnet}.  Since both points or voxels are a mono-modality input, it is difficult to infer an accurate mapping between an incomplete point cloud with a large-scale incompleteness and a complete point cloud. Therefore, those methods may perform well only on specific categories of objects or the shapes with a small-scale incompleteness. The work leveraging auxiliary data to supplement the missing information of the input point cloud for the enhancement task has rarely been studied.

\section{Methods}
\subsection{Overview}
\textbf{Problem Definition.}
The proposed view-guided completion solution is based on an assumption that the input image contains the necessary structural information of the missing shape part. 
Our goal is to recover a 3D shape $\mathcal{S}$ consisting of two parts, i.e., $\mathcal{S} = \{\mathcal{\hat{S}}_{0}, \Delta \mathcal{S}\}$, {where $\mathcal{\hat{S}}_0$ is a shape part that is close to the input partial shape  $\mathcal{S}_0$ (i.e., the difference between $\mathcal{\hat{S}}_0$ and $\mathcal{{S}}_0$ is limited), and $\Delta \mathcal{S}$ is the unknown missing shape part.}
Formally, we denote $\mathcal{M}(\mathcal{S})$ as the representation of shape $\mathcal{S}$ in modality $\mathcal{M}$, \ie $\mathcal{M} = \mathcal{P}$ for point cloud and $\mathcal{M} = \mathcal{I}$ for image.
The input of the task could be formulated as:
\begin{itemize}
\item {The partial point cloud $\mathcal{P}(\mathcal{S}_0)$.}
\item The single-view image $\mathcal{I}(\mathcal{S}_{I})$, where $\mathcal{S}_I$ refers to the observed shape from the view of image. 
\end{itemize}

\textbf{Pipeline.} 
The proposed three-stage framework is shown in Figure~\ref{fig:pipeline}. 
The first stage is used to address the cross-modality fusion problem. It maps $\mathcal{I}$ to a coarse representation of point cloud in $\mathcal{P}$. 
The second stage generates a coarse point cloud and the last stage enhances it and produce a higher-quality completed point cloud. These two stages work together to perform the cross-level fusion. 

Specifically, the first stage termed as \textit{Modality Transfer} maps $\mathcal{I}$ to a coarse representation of point cloud in $\mathcal{P}$ and then aligns the reconstructed
point cloud to the input partial shape $\mathcal{S}_0$ in 3D space. 
Then the second stage termed as \textit{Part Filter} generates a
coarse point cloud from the two aligned point cloud. Additionally, this stage also 
makes a distinction between the points which are mainly from the input partial point cloud
and the other points which are mainly reconstructed from input image $\mathcal{I}(\mathcal{S}_{I})$. 
We term the former as the \textit{Fine Part} and the latter as the \textit{Coarse Part}.  
In general, the Fine part has much higher shape quality than the Coarse part and needs only a light refinement. 
Lastly, the third stage termed as \textit{Part Refinement} takes as input both the Coarse and Fine parts and produces a completed higher-quality point cloud.
In this stage, we mainly refine the shape of the Coarse but use the Fine part as the constraint for the refinement. This can achieve better results than a global refinement which refines all the points without any constraint as confirmed by the ablation study in the experimental section. We achieve this by a novel neural network called ``Dynamic Offset Predictor". Next, we detail the above three stages.


\begin{figure}[t]
\begin{center}
  \includegraphics[width=1\linewidth]{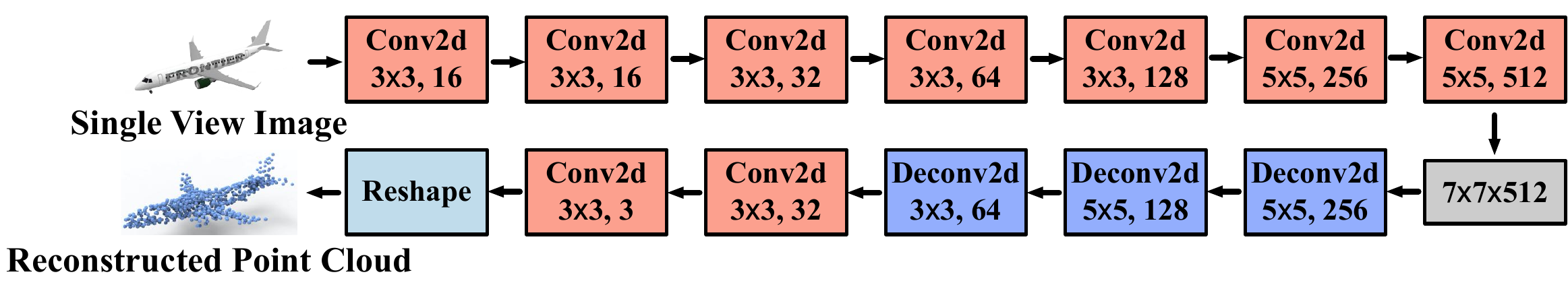}
\end{center}
\vspace{-0.3cm}
   \caption{Network structure for point cloud generation from a single view image in Modality Transfer.} 
\label{fig:PCGStructure}
\vspace{-0.5cm}
\end{figure}
\vspace{-0.2cm}
\subsection{Modality Transfer}
\label{Modality Transfer}
It would be challenging to directly reconstruct a high-quality point cloud which 
preserves rich local details from the image $\mathcal{I}(\mathcal{S}_I)$.
Thus, the transferred point cloud is expected to work as an initialized
shape mainly used for structure guidance. Let the point cloud reconstructed from the image $\mathcal{I}(\mathcal{S}_I)$ as $\mathcal{P}_{r}(\mathcal{S}_I)$. A light-weight point cloud reconstruction network with {an encoder-decoder architecture} shown in Figure~\ref{fig:PCGStructure} is employed. The encoder maps the input image into a latent space vector, and the decoder outputs an $N_{r} \times 3$ matrix, each row of which represents the Cartesian coordinates of a point. 
In our implementation, the encoder comprises a series of convolutional layers with ReLU activation, and outputs feature map in $7 \times 7 \times 512$ as the latent space vector.
The decoder applies a series of deconvolutional layers and flatten the output to generate $N_{r}$ point coordinates. 
Feature maps from each layer of the encoder are also preserved as additional guidance for the following processes in this stage.
After reconstructing $\mathcal{P}_{r}(\mathcal{S}_I)$ from the single-view image, we align it with the input partial point cloud. 


\subsection{Part Filter}
The reconstructed $\mathcal{P}_{r}(\mathcal{S}_I)$ can roughly describe the view-observed shape $\mathcal{S}_{I}$, which is assumed to contain the major information of the missing shape part $\Delta \mathcal{S}$. 
In this stage, we firstly merge $\mathcal{P}(\mathcal{S}_0)$ and $\mathcal{P}_{r}(\mathcal{S}_I)$ 
and extract a subset ($N_{c}$ points) of it as a coarse representation of shape $\mathcal{S}_0 \cup \mathcal{S}_I$. Since point density may differ in $\mathcal{P}(\mathcal{S}_0)$ and the reconstructed $\mathcal{P}_r(\mathcal{S}_I)$ and the concatenation leads to redundant points in the overlap space between $\mathcal{S}_0$ and $\mathcal{S}_I$. 
We would like this coarse point cloud with $N_{c}$ points to be uniformly dense
and preserve the global structure but also as many local details of the shape as possible.
Specifically, we use the farthest point sampling (FPS) \cite{moenning2003fast}) on $\mathcal{P}_{c}(\mathcal{S}_0 \cup \mathcal{S}_I)$ to achieve this goal.

After that, we identify the points that do not need heavy refinement
as the Fine part and leave the remaining points as the Coarse part, i.e., dividing the coarse point cloud with $N_{c}$ points into
the Coarse part ($N_{m}$ points) and Fine part ($N_{c}-N_{m}$ points). 
To search $N_{c}-N_{m}$ points that are close to the points in $\mathcal{P}(\mathcal{S}_0)$ for the Fine part, we construct the correspondence between points in $\mathcal{P}(\mathcal{S}_0)$ and $\mathcal{P}_c(\mathcal{S}_0 \cup \mathcal{S}_I)$ by Chamfer Distance (CD) \cite{fan2017point}. 
For each point $\mathbf{p}$ in $\mathcal{P}_c(\mathcal{S}_0 \cup \mathcal{S}_I)$, we calculate its distance to the closest point in $\mathcal{P}(\mathcal{S}_0)$, \ie $d(\mathbf{p}) = \min_{\mathbf{q} \in \mathcal{P}(\mathcal{S}_0)}\|\mathbf{p} - \mathbf{q}\|_2^2$. 
If the distance $d(\mathbf{p})$ is less than an adaptive threshold $d_{thr}$, we select it as a candidate point in the 
the Fine part. 
In the implementation, the threshold $d_{thr}$ can adapt to the density varying of point cloud. Technically, we randomly divide the points from $\mathcal{P}_c(\mathcal{S}_0 \cup \mathcal{S}_I)$ into two subsets, and calculate the closest distance $d(\mathbf{p})$ of all points between the two subsets. Then the average distance value could be an estimation of the point cloud density and it serves as the threshold $d_{thr}$.

\subsection{Part Refinement}

\begin{figure}[t]
\begin{center}
  \includegraphics[width=1\linewidth]{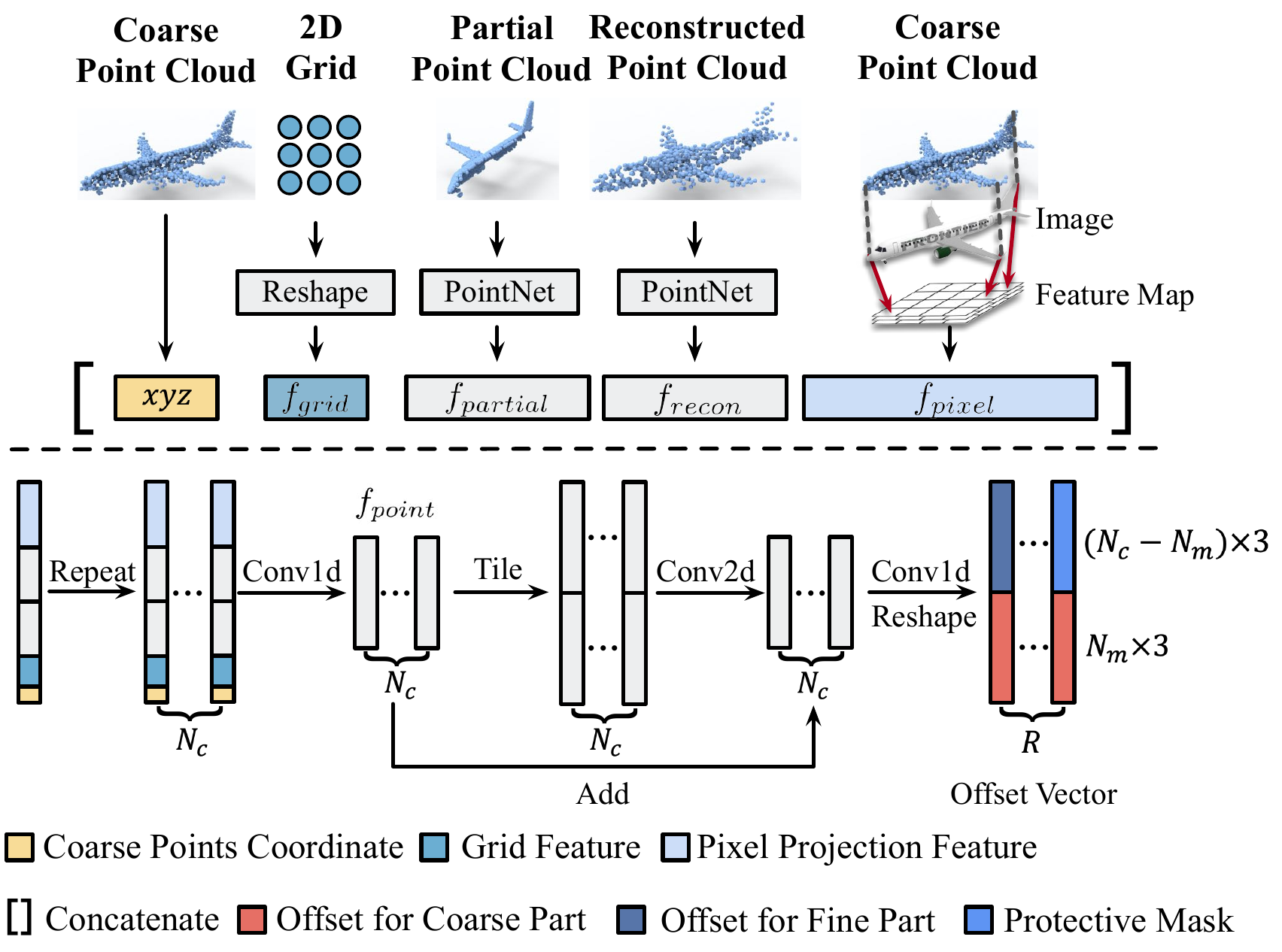}
\end{center}
   \caption{We concatenate five types of features (top row) as the input of Dynamic Offset Predictor whose architecture is shown in the bottom row.} 
\label{fig:refinement}
\vspace{-0.5cm}
\end{figure}

Part Refinement stage further refines (up-samples) the coarse point cloud consisting of the Fine and Coarse parts to produce a complete point cloud.
To achieve an effective refinement, besides of the coarse point cloud itself (represented by the point coordinates $[x,y,z]$), the refinement 
also leverages the guidance of four other types of features which can be obtained from the stages of Modality Transfer and Part Filter as shown in Figure~\ref{fig:refinement}. Those four types of features which can be classified into 2D- or 3D- guidance, together with 
the point coordinates $[x,y,z]$ of the coarse point cloud, are concatenated into a global feature vector shared by each points in the coarse point cloud. 
The proposed Dynamic Offset Predictor network takes $N_c$ repeated global feature vectors
as input and predicts the coordinate offset value for $N_c \times R $ ($R$ defines the up-sampling rate) output points. 

\textbf{3D Guidance.} 
In recent learning-based completion methods, it has been verified that completion based on the existing shape and prior knowledge is a feasible way. For example, the symmetry of airplane could help us to complete the missing wing in one side. Thus, we extract global features by applying the common encoder PointNet \cite{qi2017pointnet} on the partial point cloud $\mathcal{P}(\mathcal{S}_0)$ and the reconstructed $\mathcal{P}_r(\mathcal{S}_I)$.


\textbf{2D Guidance.}
The single-view image $\mathcal{I}(\mathcal{S}_I)$ serves the guidance for the recovery of both the structure and geometrical details of the Coarse part. 
During the stage of modality transfer, the feature of the image $\mathcal{I}(\mathcal{S}_I)$ has been extracted into feature maps with different sizes, \eg $56 \times 56$, $28 \times 28$, $14 \times 14$, $7 \times 7$. 
Inspired by the perceptual feature pooling in Pixel2Mesh \cite{wang2018pixel2mesh}, for each point, we search the features in the image feature maps corresponding to each point in the coarse point cloud using the coordinates and camera parameters. These features are stacked as $f_{pixel}$ for a kind of guidance feature. 
In addition, to increase variations among the local points and prevent predicting the same offset for different points, inspired by FoldingNet~\cite{yang2018foldingnet}, we generate a 2D-grid and reshape to a feature vector as $f_{grid}$ to boost slight disturbance. 



\vspace{-0.5cm}
\paragraph{Dynamic Offset Predictor.}
\label{Dynamic Offset Predictor}
Unlike the previous folding-based or tree structure methods, Dynamic Offset Predictor predicts the spatial offset of each point towards its current position rather than directly predicting the coordinates, which simplifies the regression. 
The data flow of Dynamic Offset Predictor shown in Figure~\ref{fig:refinement} is as follows:
\textbf{1.} The $f_{fusion}$ is repeated $N_{c}$ times and fed into a series of 1D convolutional layers to output the hidden embeddings $f_{point}$.
\textbf{2.} The $f_{point}$ is tiled $R$ times for point movements and up-sampling.
\textbf{3.} The 2D convolutional layers with kernel size $1 \times R$ are applied to perceive the local information of each point with $R$ offsets. 
\textbf{4.} The offset vector in $R \times N_{c} \times 3$ is predicted by the 1D convolutional layers.
\textbf{5.} A protective mask in $1 \times (N_{c}-N_{m}) \times 3$ with a small offset value $\epsilon$ is generated to cover the offset vector, aiming to limit the movement of Fine Part's points. The coordinates of $\mathcal{P}_{coarse}$ are then stacked $R$ times ($R \times N_{c}$) and added with corresponding offsets as the completed point cloud. Therefore, \textit{Part Refinement} with \textit{Dynamic Offset Predictor} refines and up-samples $\mathcal{P}_{coarse}$ without changing the Fine Part's detail.

\subsection{Loss Function}
The loss function measures the difference between dense point cloud and ground truth. Since the point cloud are unordered data, the loss function must be permutation invariant, Chamfer Distance (CD) and Earth Mover's Distance (EMD) is considered in this work. 
Given two subsets $P\subseteq \mathbb{R}^{3}$ and $Q\subseteq \mathbb{R}^{3}$, Chamfer distance calculates the average closest point distance between $P$ and $Q$. We use the symmetric version of CD as formulated in (\ref{equation_cd}), where the first term forces the output point cloud to move close to ground truth, and the second term ensures that the output point cloud covers the ground truth point cloud.
\begin{equation}
\mathcal{L}_{CD}=\frac{1}{\left|P\right|}\sum_{p\in P}\min_{q\in Q}\|p-q\|_{2}^{2}+\frac{1}{\left|Q\right|}\sum_{q\in Q}\min_{p\in P}\|p-q\|_{2}^{2}
\label{equation_cd}
\end{equation}

The Earth Mover's Distance (EMD) is an algorithm to evaluate the dissimilarity between two multi-dimensional distributions.
Let $\phi:P\rightarrow Q$ be a bi-jection between two point clouds, which finds the minimal average distance between two point sets. In practice, searching the optimal $\phi$ is computationally expensive, we thereby use an iterative $(1+\epsilon)$ approximation scheme as ~\cite{bertsekas1985distributed}. 
\begin{equation}
\mathcal{L}_{EMD}=\min_{\phi:P\rightarrow Q}\frac{1}{\left|P\right|}\sum_{p\in P}\|p-\phi(p)\|_{2}^{2}
\label{equation_emd}
\end{equation}
We combine these two distance with $ \mathcal{L} = \alpha\mathcal{L}_{CD} + \beta\mathcal{L}_{EMD} $
as the loss function, where  $\alpha$, $\beta$ are the trade-off hyper-parameters. We use $\alpha=1$ and $\beta=0.0001$ as a default value in the training stage.


\section{Experimental Settings}

\begin{table*}[]
\begin{center}
\setlength{\tabcolsep}{3.3mm}{
\renewcommand{\arraystretch}{1.2}
\begin{tabular}{l|c|c|c|c|c|c|c|c|c}
\toprule
\multirow{2}{*}{Methods} & \multicolumn{9}{c}{Mean Chamfer Distance per point} \\ \cline{2-10} 
 & Avg & Airplane & Cabinet & Car & Chair & Lamp & Sofa & Table & Watercraft \\ \toprule \hline
AtlasNet \cite{groueix2018} & 6.062 & 5.032 & 6.414 & 4.868 & 8.161 & 7.182 & 6.023 & 6.561 & 4.261 \\ 
FoldingNet \cite{yang2018foldingnet} & 6.271 & 5.242 & 6.958 & 5.307 & 8.823 & 6.504 & 6.368 & 7.080 & 3.882 \\ 
PCN \cite{yuan2018pcn} & 5.619 & 4.246 & 6.409 & 4.840 & 7.441 & 6.331 & 5.668 & 6.508 & 3.510 \\ 
TopNet \cite{tchapmi2019topnet} & 4.976 & 3.710 & 5.629 & 4.530 & 6.391 & 5.547 & 5.281 & 5.381 & 3.350 \\ 
Ours & \textbf{3.308} & \textbf{1.760} & \textbf{4.558} & \textbf{3.138} & \textbf{2.476} & \textbf{2.867} & \textbf{4.481} & \textbf{4.990} & \textbf{2.197} \\ \bottomrule
\end{tabular}}
\end{center}
\caption{Quantitative results on ShapNet-ViPC using Chamfer Distance with 2,048 points. The best results are highlighted in bold.}
\vspace{-0.1cm}
\label{table1:cdTable}

\end{table*}

\begin{table*}[]
\begin{center}
\setlength{\tabcolsep}{3.3mm}{
\renewcommand{\arraystretch}{1.2}
\begin{tabular}{l|c|c|c|c|c|c|c|c|c}
\toprule
\multirow{2}{*}{Methods} & \multicolumn{9}{c}{F-Score@0.001} \\ \cline{2-10} 
 & Avg & Airplane & Cabinet & Car & Chair & Lamp & Sofa & Table & Watercraft \\ \toprule \hline
AtlasNet \cite{groueix2018} & 0.410 & 0.509 & 0.304 & 0.379 & 0.326 & 0.426 & 0.318 & 0.469 & 0.551 \\ 
FoldingNet \cite{yang2018foldingnet} & 0.331 & 0.432 & 0.237 & 0.300 & 0.204 & 0.360 & 0.249 & 0.351 & 0.518 \\ 
PCN \cite{yuan2018pcn} & 0.407 & 0.578 & 0.270 & 0.331 & 0.323 & 0.456 & 0.293 & 0.431 & 0.577 \\ 
TopNet \cite{tchapmi2019topnet} & 0.467 & 0.593 & 0.358 & 0.405 & 0.388 & 0.491 & 0.361 & 0.528 & 0.615 \\ 
Ours & \textbf{0.591} & \textbf{0.803} & \textbf{0.451} & \textbf{0.5118} & \textbf{0.529} & \textbf{0.706} & \textbf{0.434} & \textbf{0.594} & \textbf{0.730} \\ \bottomrule
\end{tabular}}

\end{center}
\caption{Quantitative results on ShapNet-ViPC using F-Score with 2,048 points. The best results are highlighted in bold.}
\label{table2:fscoreTable}
\vspace{-0.3cm}
\end{table*}

\subsection{ShapeNet-ViPC}
\begin{figure}[t]
\begin{center}
  \includegraphics[width=1\linewidth]{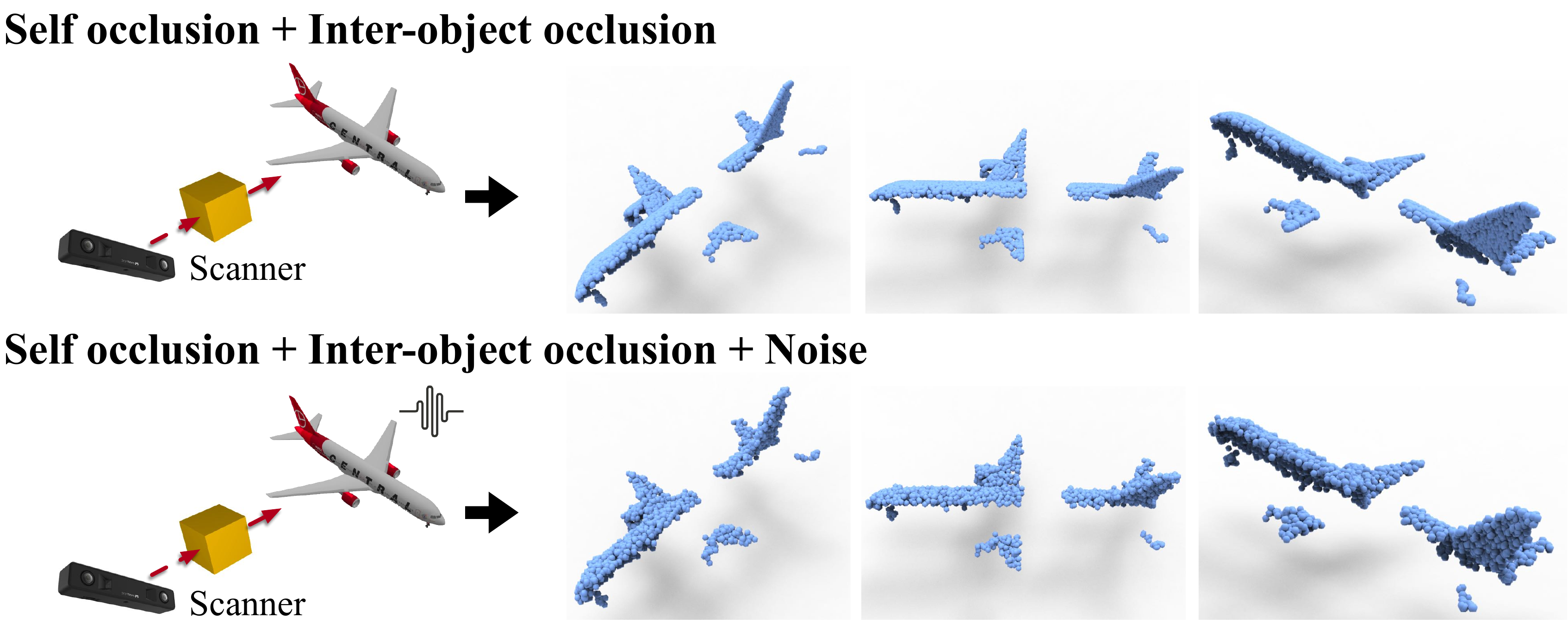}
\end{center}
   \caption{Illustration of two typical types of point cloud acquisition scenarios.
   Top: the target shape is occluded by other objects in the environment as well as a part of itself (self-occlusion); bottom: besides of self- and inter-object occlusion, the locations of the acquired points are disturbed because of device noise.}
\label{fig:data_prepare}
\vspace{-0.5cm}
\end{figure}

To simulate the defects related to the task of ViPC and evaluate the performance of the proposed approach, we build a new dataset called ShapeNet-ViPC based on the ShapeNetRendering\cite{chang2015shapenet}.
It contains 38,328 objects from 13 categories, \ie, airplane, bench, cabinet, car, chair, monitor, lamp, speaker, firearm, sofa, table, cellphone, watercraft. 
For each object, we generate two types incomplete point cloud (with or without noise) under 24 view points, as illustrated in Figure~\ref{fig:data_prepare}. 
The 24 view points follow the same view point setting as ShapeNetRendering   \cite{chang2015shapenet} (this setting is also used in 3D-R2N2 \cite{choy20163d}, PointSetGeneration \cite{fan2017point} and Pixel2Mesh \cite{wang2018pixel2mesh}). 
For each set we uniformly sample 2,048 points from the mesh surface of a target shape under the corresponding view point setting as the ground-truth complete point cloud. Specifically, each 3D shape is normalized into the bounding sphere with radius of 1, 
and rotated to the pose corresponding to a specific view point lastly. For the image data, we use the same 24 rendered views as ShapeNetRendering. 
ShapeNet-ViPC contains $38,328 \times 24 = 919,872$ sets of training data in total.
Each set contains one ground-truth complete point cloud, two incomplete (partial) point clouds, and an image view. In this paper, we use 31,650 objects (759,600 sets) of eight categories for all experiments, 80\% for training and 20\% for test.

\subsection{Implementation details and evaluation metrics}
In our implementation, the size of input image is $224 \times 224$, from which
we reconstruct a point cloud with $N_m = 784$ points. The partial input contains 2,048 points. We sample a coarse point cloud with $N_c = 1024$ points by  FPS~\cite{moenning2003fast} from the combination of the input partial point cloud and the reconstructed point cloud (i.e., $\mathcal{P}_c(\mathcal{S}_0 \cup \mathcal{S}_I)$). The output complete cloud contains 2,048 points (i.e.,$R = 2 $). The network for modality transfer is pre-trained with batch size of 64 and learning rate of 1e-4 for 100 epochs. The Part Refinement network is trained with batch size of 1 and learning rate 1e-6 for 200 epochs.
Category-specific parameters are trained for each object category.
To quantify the completion performance, we use both the Chamfer Distance (CD)~\cite{tatarchenko2019single} and F-Score~\cite{tatarchenko2019single} as the quantitative evaluation metrics. Results with lower CD value and/or higher F-Score correspond to better completion quality.

\section{Experimental Results and Analysis}

\begin{figure*}[t]
\begin{center}
  \includegraphics[width=1\linewidth]{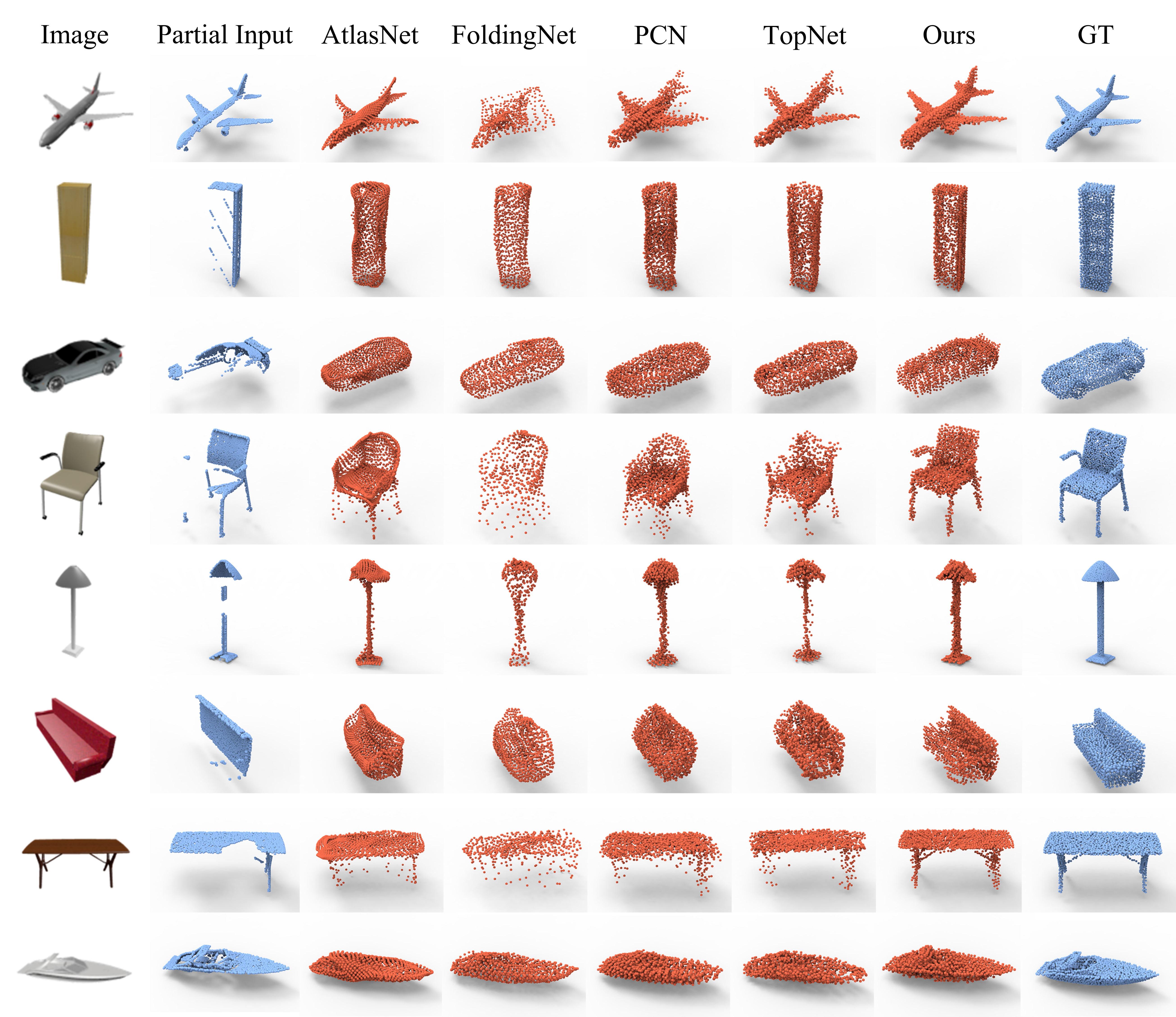}
\end{center}
   \caption{Qualitative comparison on ShapeNet-ViPC. Our method outperforms other baseline methods with significant margins. The resolution for partial, completed and groud truth point clouds are 2,048.}
\label{fig:compare}
\vspace{-0.3cm}
\end{figure*}

\subsection{Comparisons}
We compare our method with several state-of-the-art methods on the task of point cloud completion, including AtlasNet \cite{groueix2018}, FoldingNet \cite{yang2018foldingnet}, point completion network (PCN) \cite{yuan2018pcn} and TopNet \cite{tchapmi2019topnet}. AtlasNet recovers a complete point cloud by estimating a collection of parametric surface elements. FoldingNet, 2D-grid based auto-encoder, is a pioneer grid based method for point cloud completion. PCN is an encoder-decoder framework which completes a partial input point cloud with a typical coarse-to-fine scheme. TopNet completes an imperfect point cloud by a tree structure network. All the above baseline methods take as input only a partial point cloud.

\textbf{Quantitative results}. 
We normalize the output point clouds produced by the comparison methods
and calculate the CD and F-Score on the 2,048 points of each shape.
The results on each categories and the average are summarized in Tables \ref{table1:cdTable} and \ref{table2:fscoreTable}. It is found that the proposed method consistently outperforms other methods with a
significant margin on all the eight categories on both CD and F-Score metrics. 
Besides, our method demonstrates more advantages on the categories of airplane, watercraft, lamp, and car.

\textbf{Qualitative results}. 
We also visualize the results produced by the comparison methods for a more comprehensive evaluation. Results on the representative examples from the eight categories are shown in Figure~\ref{fig:compare}. It is easy to observe that the completed point clouds produced by FoldingNet are relatively messy. The generated point clouds do not show clear structures on some shape parts, e.g., the wings of the airplane, the chair legs, the table legs. PCN and AtlasNet produce improved qualitative results compared to FoldingNet overall. However, local small-scale structural details are still missing (e.g., the fuel tank of the airplanes, the arms of the chairs) in
the results. The structured-tree based TopNet method achieves better visual results than PCN and AtlasNet in general. We can see the evidences on the airplane, lamp, sofa, and watercraft, which exhibits much clearer part structures and points arranged more neatly. However, some part details in the input partial point clouds are not preserved in the completed point clouds, e.g., the points of the fuel tank of the airplane and the trestle of the table have been moved to other parts. Results produced by our method have
no this problem and show visually better performance on all eight categories than baselines. This is because unlike other comparison methods inferring the locations of all points, our method instead uses the points in the other part of the shape (i.e., Fine part) as a completion constraint and infers a local point distribution of a part of the shape (i.e., Coarse part). This makes the network preserve the local detail from the input partial point cloud, produce a more reasonable completion for the missing structure (see the reconstructed left trestles of the table), as well as converge much faster.


\vspace{0.1cm}
\subsection{Experimental Analysis}
\textbf{Ablation Studies.}
We conduct ablation experiments to study the individual contributions of each stage in our method. 
As shown in Table~\ref{table3:ablationTable}, we quantitatively compare the point cloud qualities of the reconstructed point clouds $P_{rec}$ generated by Modality Transfer, the coarse point clouds $P_{coarse}$ generated by Part Filter, and the completed point clouds $P_{complete}$ generated by Part Refinement. 
In addition, we also study how much the {differential refinement strategy} that differentiates between the Fine and {Coarse parts} contributes. Specifically, we modify the Dynamic Offset Predictor network and disable the constraint that the points in the Fine part should be moved within a small sphere, i.e., the points from both the Fine and Coarse parts are processed without any difference. We term the results produced by this architecture as $P_{global}$. By comparing with $P_{global}$ and $P_{complete}$, we can obtain the quantitative contribution of the differential refinement strategy.


\begin{figure}[t]
\begin{center}
  \includegraphics[width=1.05\linewidth]{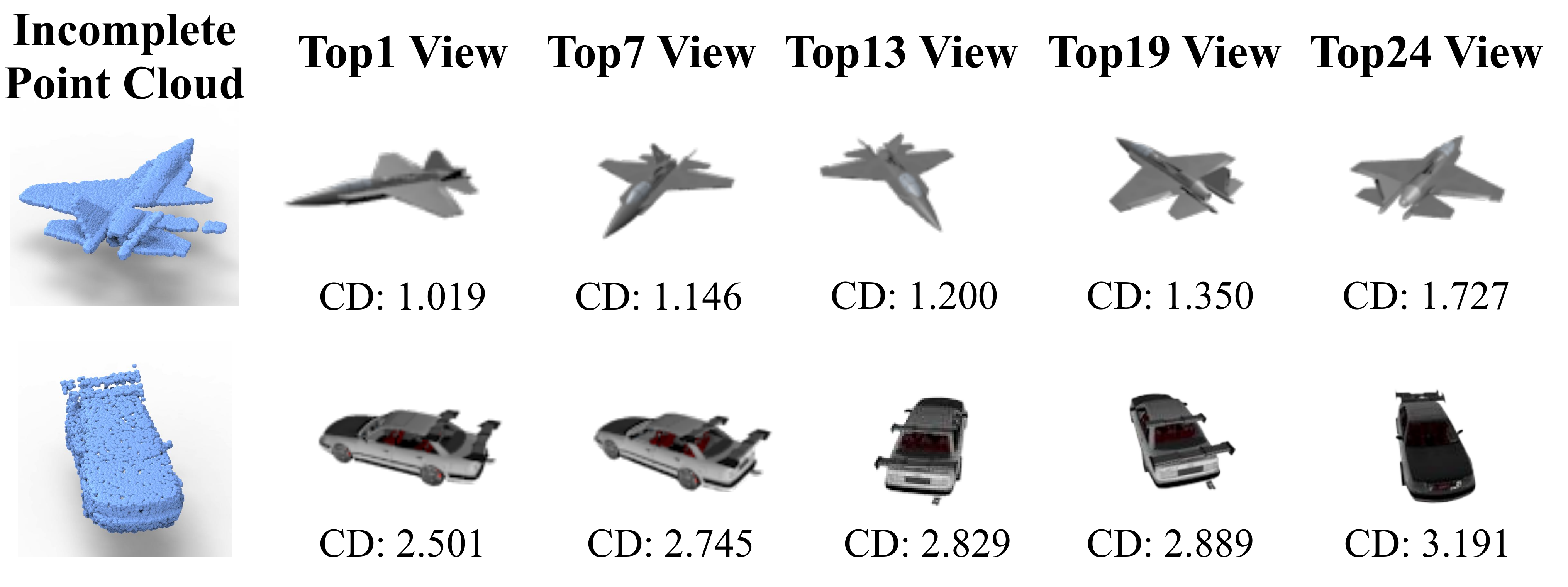}
\end{center}
   \caption{Views provides more complementary information for the input partial point clouds can produce better completion results. Each input partial point clouds are shown on the left; quantitative completion performance measured by the average CD (unit: $10^{-3}$) is reported below each input reference view.}
\label{fig:exp_view}
\vspace{-1cm}
\end{figure}

\textbf{Contribution of the Single-view Image.}
In this set of experiments, we study what kind of input view can better improve the completion. We randomly select $50 \times 8 = 400$ partial point clouds (50 partial point clouds for each category) from the test set of ShapeNet-ViPC for the evaluation. For each partial point cloud, we produce 24 complete point clouds, each of which is generated with the reference of an image from the 24 rendered views. We quantify the completion quality of those 400 completion point clouds with the CD metric and demonstrate some representative results
in Figure~\ref{fig:exp_view}. It indicates that different image views can provide different improvement.
The image views which can provide more information for the missing part of the partial point would produce better results. 


\begin{table}[]
\begin{center}
\setlength{\tabcolsep}{2.35mm}{
\begin{tabular}{c|c|c|c|c}
\toprule
\multirow{2}{*}{Category} & \multicolumn{4}{c}{Mean Chamfer Distance per point} \\ \cline{2-5} 
                         & $P_{rec}$   & $P_{coarse}$   & \tabincell{c}{$P_{global}$ }  & \tabincell{c}{$P_{complete}$ }  \\ \hline \toprule \hline
Airplane                 & 4.479 &2.360& 1.993 & \textbf{1.760}\\
Cabinet                  & 7.381  &5.531& 4.807 & \textbf{4.558}\\
Car                      & 4.975  &3.921& 3.308 & \textbf{3.138}\\
Chair                    & 12.198  &6.967& 6.098 & \textbf{2.476}\\
Lamp                     & 4.573 &3.549& 3.186 & \textbf{2.867}\\
Sofa                     & 7.809  &5.340& 4.681 & \textbf{4.481}\\
Table                    & 10.967  &6.719& 5.891 & \textbf{4.990}\\
Watercraft               & 5.626  &3.156& 2.669 & \textbf{2.197}\\ \hline
mean                     & 7.241 & 4.693& 4.079 & \textbf{3.308}\\ 
\bottomrule
\end{tabular}}
\vspace{0.1cm}
\caption{Quantitative results for the ablation study; average CD (unit: $10^{-3}$) are reported.}
\label{table3:ablationTable}
\end{center}
\vspace{-0.3cm}
\end{table}

\begin{figure}[t]
\begin{center}
  \includegraphics[width=1.0\linewidth]{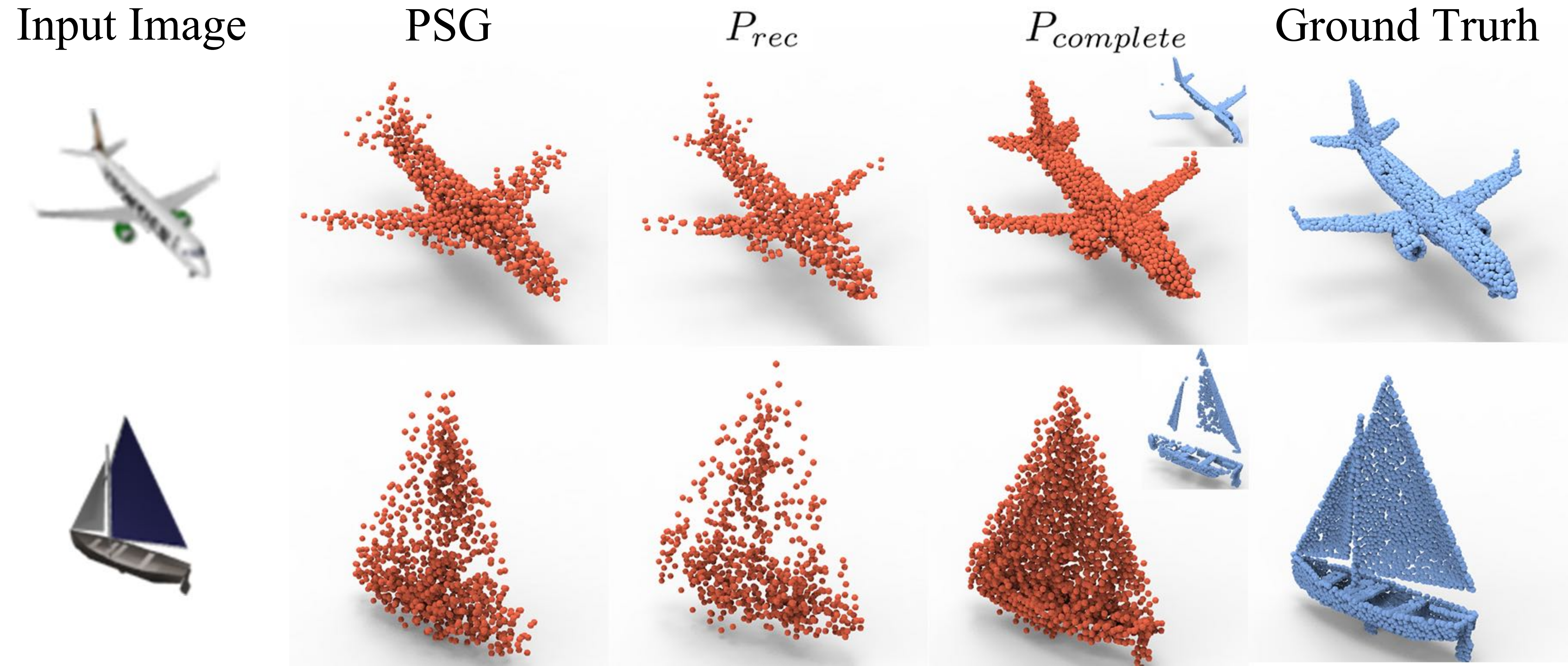}
\end{center}
   \caption{Qualitative comparison with PSG. The input views are shown on the left; extra input partial point clouds for $P_{complete}$ are shown in the top right corner of each corresponding results.}
\label{fig:compare_to_psg}
\vspace{-0.5cm}
\end{figure}

\textbf{Contrast to SVR Methods.}
In this set of experiments, we study if the proposed ViPC is superior to
the SToA single-view based reconstruction method PSG~\cite{fan2017point}.
We compare three different architectures: PSG, the point generation network in the Modality Transfer stage (i.e., $P_{rec}$ in Table~\ref{table3:ablationTable}), and the entire proposed framework (i.e., $P_{complete}$ in Table~\ref{table3:ablationTable} on the test set of ShapeNet-ViPC. We compute the average CD value of the results produced by the three comparison architectures. 
We summarize the quantitative results in Table~\ref{table4:SVR_table} and visualize representative results in Figure~\ref{fig:compare_to_psg}. Those results indicate
the entire proposed framework outperform exceed PSG with a large performance margin. We also find the results generated by PSG show better quality than $P_{ren}$. Replacing the adopted network in Modality Transfer with a more effective network like PSG may provide a better solution. 

\textbf{Feature Ablation Studies.}
We remove $xyz$, $f_{partial}$, $f_{recon}$, $f_{pixel}$, $f_{grid}$ from $f_{fusion}$, respectively, to study the contribution of different features in $f_{fusion}$ during the Part Refinement. The results of completion performance are shown in Table~\ref{table5:featureAblation}. Their contributions can be ranked as $xyz \textgreater f_{partial} \approx f_{recon} \textgreater f_{pixel} \textgreater f_{grid}$.

\begin{table}[]
\begin{center}
\setlength{\tabcolsep}{4.25mm}{
\begin{tabular}{c|c|c|c}
\hline
Methods  & PSG \cite{fan2017point}   & $P_{rec}$ & $P_{complete}$  \\ \hline
CD.($10^{-3}$) & 7.092 & 7.241 & 3.308 \\ \hline
\end{tabular}}
\vspace{0.05cm}
\caption{Quantitative comparison with PSG.}
\label{table4:SVR_table}
\end{center}
\vspace{-0.5cm}
\end{table}

\begin{table}[]
\begin{center}
\setlength{\tabcolsep}{1.1mm}{
\begin{tabular}{c|c|c|c|c|c|c}
\hline
Remove  & $xyz$   & $f_{partial}$ & $f_{recon}$ & $f_{pixel}$ & $f_{grid}$ & \textit{none}  \\ \hline
CD.($10^{-3}$) & 3.348 & 3.325 & 3.324 & 3.321 & 3.316 & 3.308 \\ \hline
Decline(\%) & 1.22 & 0.52 & 0.49 & 0.39 & 0.24 & - \\ \hline
\end{tabular}}
\vspace{0.05cm}
\caption{Quantitative results of removing different features and their relative decline compared with fusion of all features.}
\label{table5:featureAblation}
\end{center}
\vspace{-0.7cm}
\end{table}

\subsection{Limitations}
\textbf{Point cloud registration.} 
To ensure the Part Filter and Part Refinement giving full play to their effects, the input partial point cloud and the reconstructed point cloud should be aligned accurately. Because the camera parameters are not hard to obtain in calibrated devices. We align the input partial point cloud with the reconstructed point cloud by using camera parameters in Modality Transfer stage. 
As for non-calibrated devices, we tried several classical unsupervised point cloud registration methods such as ICP \cite{besl1992method} and ICP-MCC \cite{du2020robust} to perform the point cloud registration. Unfortunately, due to the sparsity of the reconstructed point clouds and the incompleteness of the partial input, it is hard to achieve an accurate
alignment. Replacing the current registration solution in Modality Transfer with
an effective learning based one is worthy of further study in the future.




\textbf{Completion in real-world scenes.} We have also evaluated the proposed method in real-world scenes where we captured the single view images by mobile phone and collected the partial point clouds by the Li-DAR device in the iPad Pro. Because we train our model on rendered images where the rendered textures can not reflect the illumination in the real environment, it leads to poor-quality reconstructed point clouds in the Modality Transfer stage and thus produces
unsatisfactory completion results. Training on real-world data may solve this problem. 
\section{Conclusion}
We propose a pioneer sensor fusion work called ViPC for the task of point cloud completion. ViPC is a view-guided point cloud completion framework. It takes the missing global structure information from an extra single-view image to complete a partial point cloud. The core technical contribution in ViPC is a point cloud refinement network called ``Dynamic Offset Predictor" which can deferentially refine the points in a coarse point cloud. We compare ViPC with existing single-modality based STOA methods that reconstruct a complete point cloud based on either the point cloud modality or the image modality. It demonstrates significant quality improvement on a new large-scale dataset we collected for the point cloud completion task.

\textbf{Acknowledgments.} 
This work was supported by National Natural Science Funds of China (U1701262, U1801\\263) and Tsinghua University Initiative Scientific Research Program (No. 20197020003).

{\small
\bibliographystyle{ieee_fullname}
\bibliography{egbib}
}

\end{document}